\documentclass[letterpaper, 10 pt, conference]{ieeeconf}  % Comment this line out if you need a4paper

\IEEEoverridecommandlockouts                              % This command is only needed if 
\makeatletter
\let\NAT@parse\undefined

\newcommand{\conf}[1]{%
\def\ps@titlepagestyle{% default title page headers, no footers
\def\@oddhead{
    \centering
    \parbox[b]{0.9\textwidth}{\vspace{-1cm} \centering \footnotesize #1}
}%
\let\@evenhead\@empty
\let\@evenfoot\@empty}
}
\makeatother

\usepackage{graphicx} % for pdf, bitmapped graphics files
\usepackage{amsmath}
\usepackage{amssymb}

\usepackage{bm}
\usepackage{bbm}
\usepackage{dsfont}
\usepackage{multirow}
\usepackage{booktabs}
\usepackage{nicefrac}
\usepackage{cite} 	% merge citations, e.g., [3]-[5]
\usepackage{pdfpages}
\usepackage[utf8]{inputenc}

\usepackage{graphicx}

\usepackage{subcaption}
\usepackage{refcount}
\usepackage{afterpage}
\newcommand{\setfootnotemark}{\refstepcounter{footnote}\footnotemark[\value{footnote}]}

\usepackage{siunitx}
\usepackage{hyperref}

\usepackage[capitalize]{cleveref} % ranges of equations

\hypersetup{
	colorlinks=true,	% remove ugly boxes
	pdfstartview=FitV,	% ???
	linkcolor=black!50!blue,	% Links within the PDF, e.g. Tables, Figures, ...
	citecolor=black!50!green,	% Links into References
	urlcolor=blue,		% Web-Links
	breaklinks=true, 	% ???
}

\newcommand*{\norm}[1]{\left\lVert#1\right\rVert}		% Norm
\newcommand*{\N}{\mathds{N}}							% natural numbers
\newcommand{\yaw}{\phi}                                 % yaw
\newcommand{\radvel}{v_r}                               % radial velocity

\conf{Copyright 2021 IEEE. Personal use of this material is permitted. Permission from IEEE must be obtained for all other uses, in any current or future media, including reprinting/republishing this material for advertising or promotional purposes, creating new collective works, for resale or redistribution to servers or lists, or reuse of any copyrighted component of this work in other works.}

\title{\LARGE \bf
The Radar Ghost Dataset -- An Evaluation of Ghost Objects in Automotive Radar Data
}

\author{% <-this % stops a space
	Florian Kraus$^{1,2}$, Nicolas Scheiner$^{1}$, Werner Ritter$^{1}$, Klaus Dietmayer$^{2}$% <-this % stops a space
	\thanks{$^{1}$Mercedes-Benz AG, 70565 Stuttgart, Germany}% <-this % stops a space
	\thanks{$^{2}$Institute of Measurement, Control and Microtechnology, Ulm University, 89081 Ulm, Germany}% <-this % stops a space
}

\begin{document}

\maketitle
\pagestyle{plain}

%%%%%%%%%%%%%%%%%%%%%%%%%%%%%%%%%%%%%%%%%%%%%%%%%%%%%%%%%%%%%%%%%%%%%%%%%%%%%%%%
\begin{abstract}
Radar sensors have a long tradition in advanced driver assistance systems (ADAS) and also play a major role in current concepts for autonomous vehicles.
Their importance is reasoned by their high robustness against meteorological effects, such as rain, snow, or fog, and the radar's ability to measure relative radial velocity differences via the Doppler effect.
The cause for these advantages, namely the large wavelength, is also one of the drawbacks of radar sensors.
Compared to camera or lidar sensor, a lot more surfaces in a typical traffic scenario appear flat relative to the radar's emitted signal.
This results in multi-path reflections or so called ghost detections in the radar signal.
Ghost objects pose a major source for potential false positive detections in a vehicle's perception pipeline.
Therefore, it is important to be able to segregate multi-path reflections from direct ones.
In this article, we present a dataset with detailed manual annotations for different kinds of ghost detections.
Moreover, two different approaches for identifying these kinds of objects are evaluated.
We hope that our dataset encourages more researchers to engage in the fields of multi-path object suppression or exploitation.
\end{abstract}

%%%%%%%%%%%%%%%%%%%%%%%%%%%%%%%%%%%%%%%%%%%%%%%%%%%%%%%%%%%%%%%%%%%%%%%%%%%%%%%%
\section{Introduction}
\addtocounter{footnote}{2} % Fix footnote counter

Doppler radar sensors are a crucial part for automated driving.
They provide a radial velocity within a single measurement combined with an unmatched robustness against adverse whether conditions.
This makes it possible to detect moving objects in almost all conditions.
The resistance against environmental effects is due to the relatively large wavelength of around \SI{4}{\mm} for a \SI{77}{\giga\hertz} radar.

\begin{figure}
	\centering
	\includegraphics[width=0.99\linewidth]{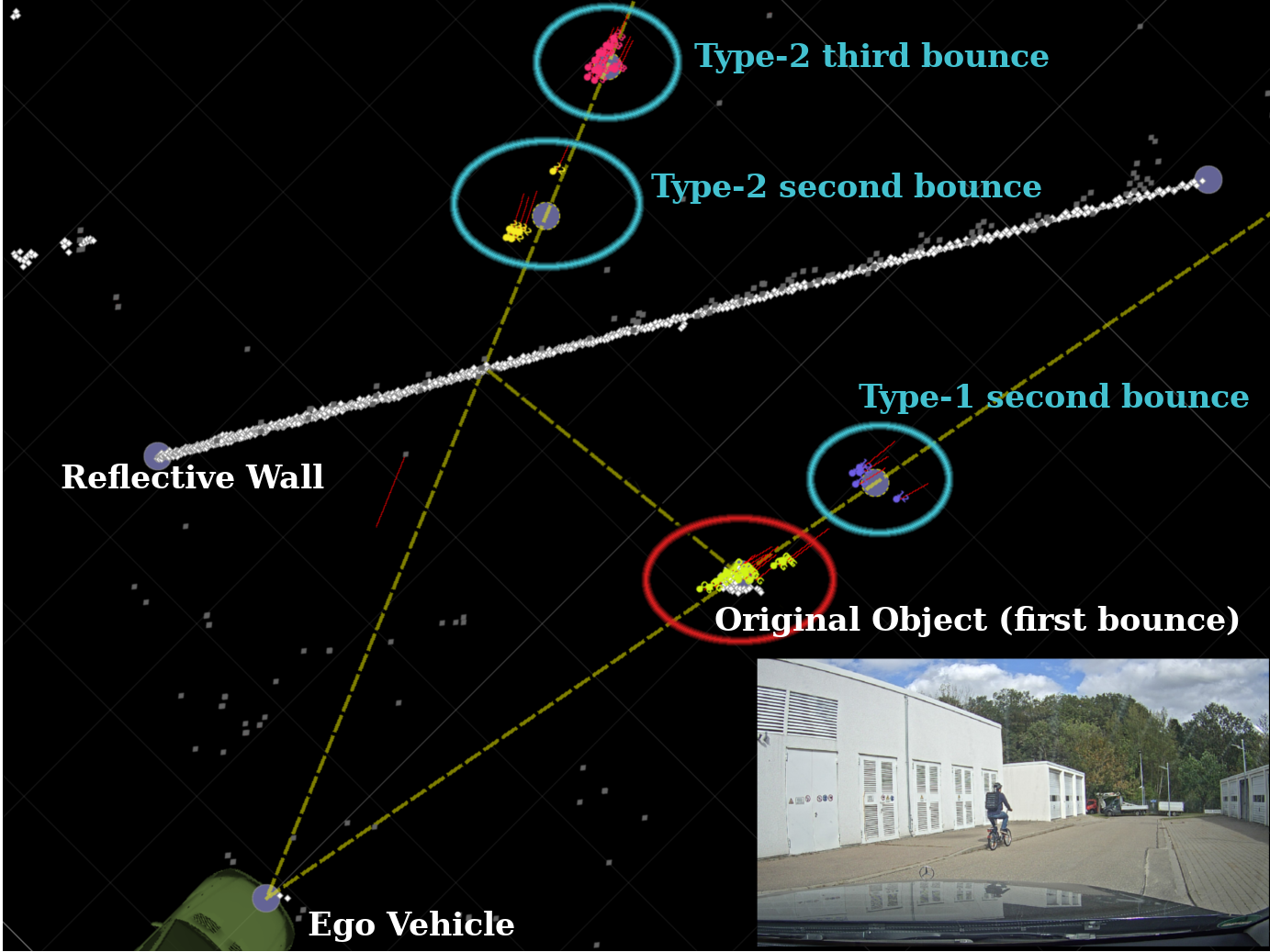}
	\caption{
		Our dataset provides detailed annotations for various types of ghost targets allowing to conduct in-depth experiments on their influence.
		Example image from our dataset.
		Different types of multi-path reflections are highlighted alongside the real object.
		The small white dots correspond to a lidar reference system, the gray ones are the radar reflections.
		See \autoref{sec:radar-ghosts} for a detailed explanation of radar ghost objects.
	}
	\label{fig:title}
	\vspace{-5mm}
\end{figure}

As a downside, many surfaces act as mirror surfaces by preserving specular radar reflections\setfootnotemark.
The caused multi-path or ghost detections can provide viable information.
For example, occluded cars following closely behind of each other can be detected utilising multi-path propagation along the road surface.
Due to the flat incidence angle which is most common for those reflections, their estimated location is almost identical to the real object position.
Multi-path reflections caused by vertical walls can easily lead to false positive detections, cf. \autoref{fig:title}.
However, they may also be exploited to detect objects around street corners using non-line-of-sight reconstruction as shown in \cite{Scheiner2020CVPR}.

In previous work \cite{Kraus2020ITSC}, we analyzed the impact of ghost object occurrences for semantic segmentation tasks on automotive radar using the data provided in \cite{Scheiner2020CVPR}.
Encouraged by the success, we completely relabeled the original dataset which was mainly annotated to reconstruct a single specific type of ghost objects.
For the proposed dataset we provide a larger number and more detailed annotations for different types of ghost objects.
The following contributions are made:
\begin{itemize}
	\item New detailed manual ghost object annotations for the original dataset in \cite{Scheiner2020CVPR} are created. This enables in-depth analysis and design of countermeasures for different types of multi-path reflections.
	\item Additional sequences are released and multi-object sequences are synthesized.
	\item A detailed analysis of the dataset and the labeling process is provided.
	\item An end to end radar object detection method based on SGPN \cite{wang2018sgpn} is presented.
	\item The impact of ghost objects on SGPN is analyzed and compared to a baseline consisting of an extended version of our previous semantic segmentation approach.
\end{itemize}

 % \label{fig:title}
\footnotetext{Specular (or regular) reflections are mirror-like reflections of waves.
	The reflected signal/ray emerges at the same angle to the surface normal as the incident signal/ray.}  % use setfootnotemark instead of \footnotemark. otherwise hyperref adds strange link to first page.

\begin{figure}
	\centering
	\begin{subfigure}{.31\columnwidth}
		\includegraphics[width=\columnwidth]{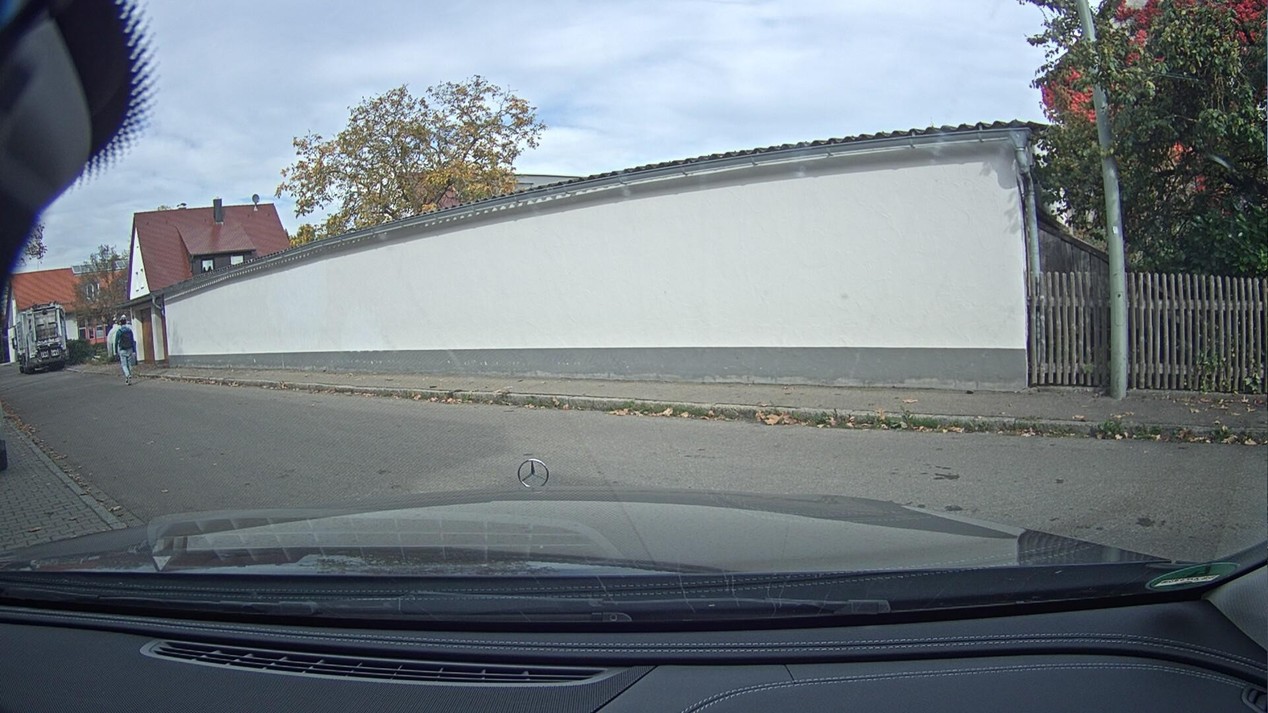}
		\captionsetup{skip=1pt} \caption*{Wall}
	\end{subfigure}
	\begin{subfigure}{.31\columnwidth}
		\includegraphics[width=\columnwidth]{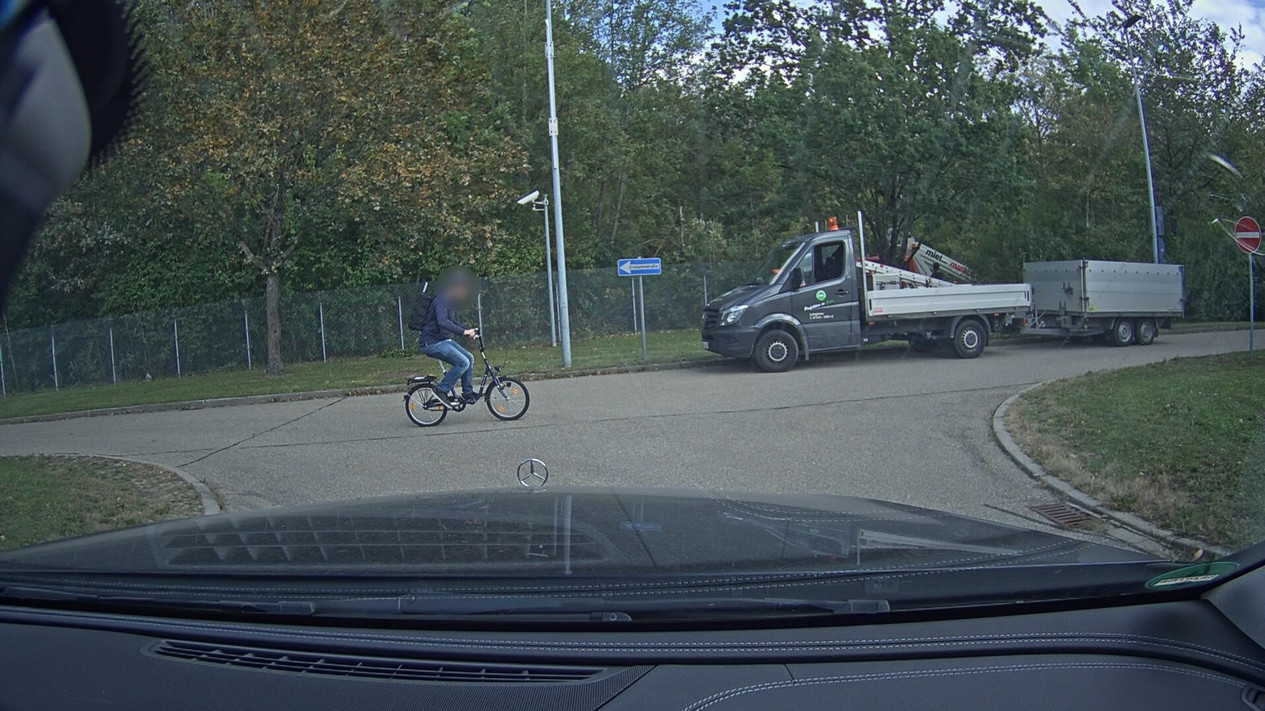}
		\captionsetup{skip=1pt} \caption*{Truck and Trailer}
	\end{subfigure}
	\begin{subfigure}{.31\columnwidth}
		\includegraphics[width=\columnwidth]{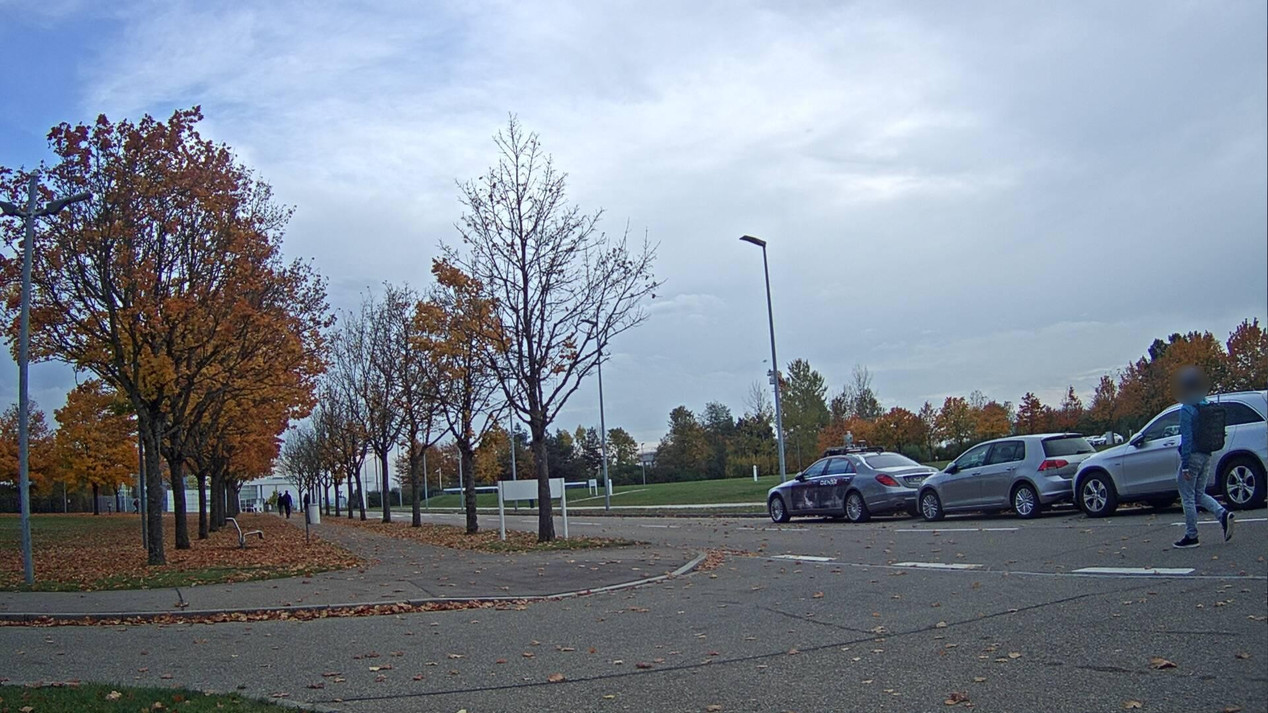}
		\captionsetup{skip=1pt} \caption*{Multiple Cars}
	\end{subfigure}

	\begin{subfigure}{.31\columnwidth}
		\includegraphics[width=\columnwidth]{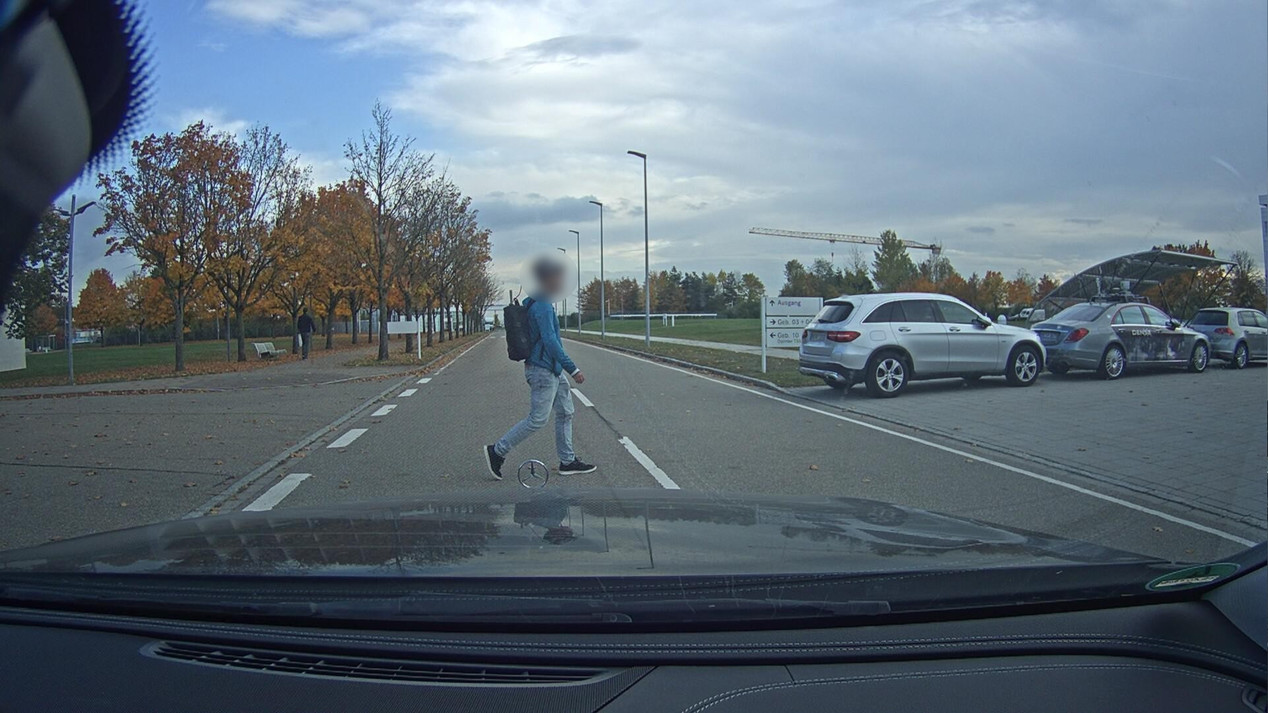}
		\captionsetup{skip=1pt} \caption*{Multiple Cars}
	\end{subfigure}
	\begin{subfigure}{.31\columnwidth}
		\includegraphics[width=\columnwidth]{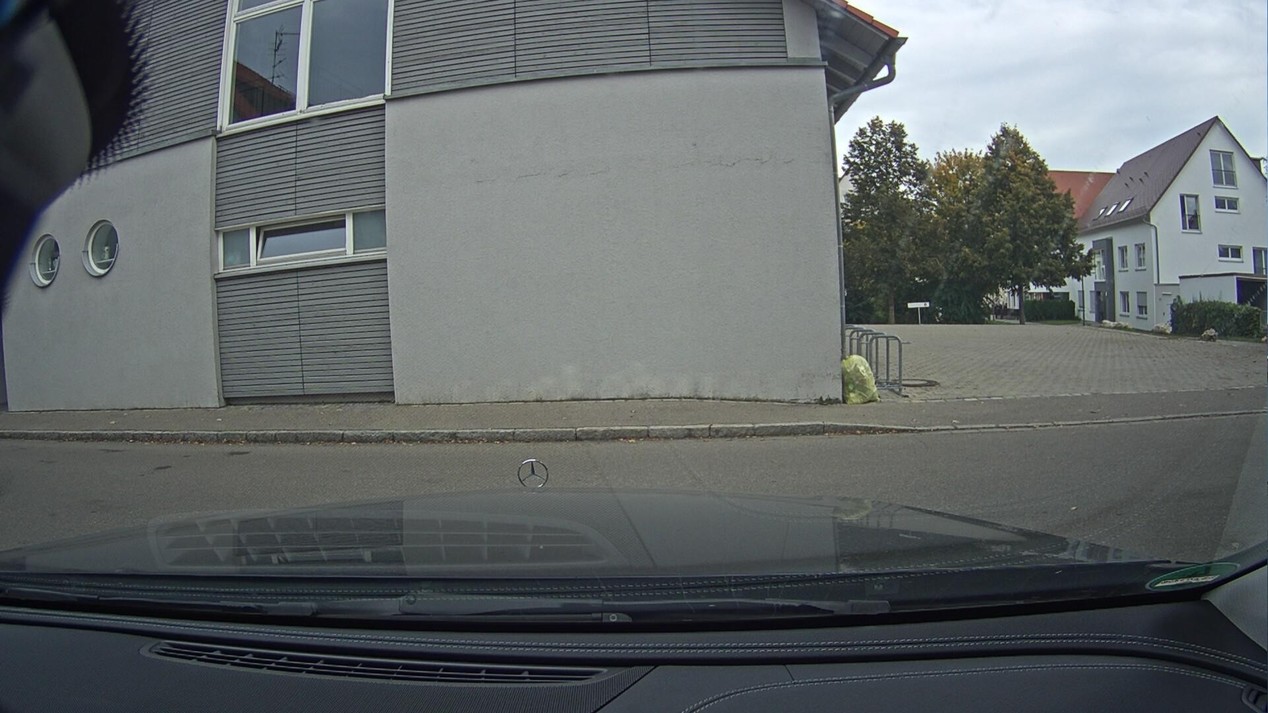}
		\captionsetup{skip=1pt} \caption*{House Facade}
	\end{subfigure}
	\begin{subfigure}{.31\columnwidth}
		\includegraphics[width=\columnwidth]{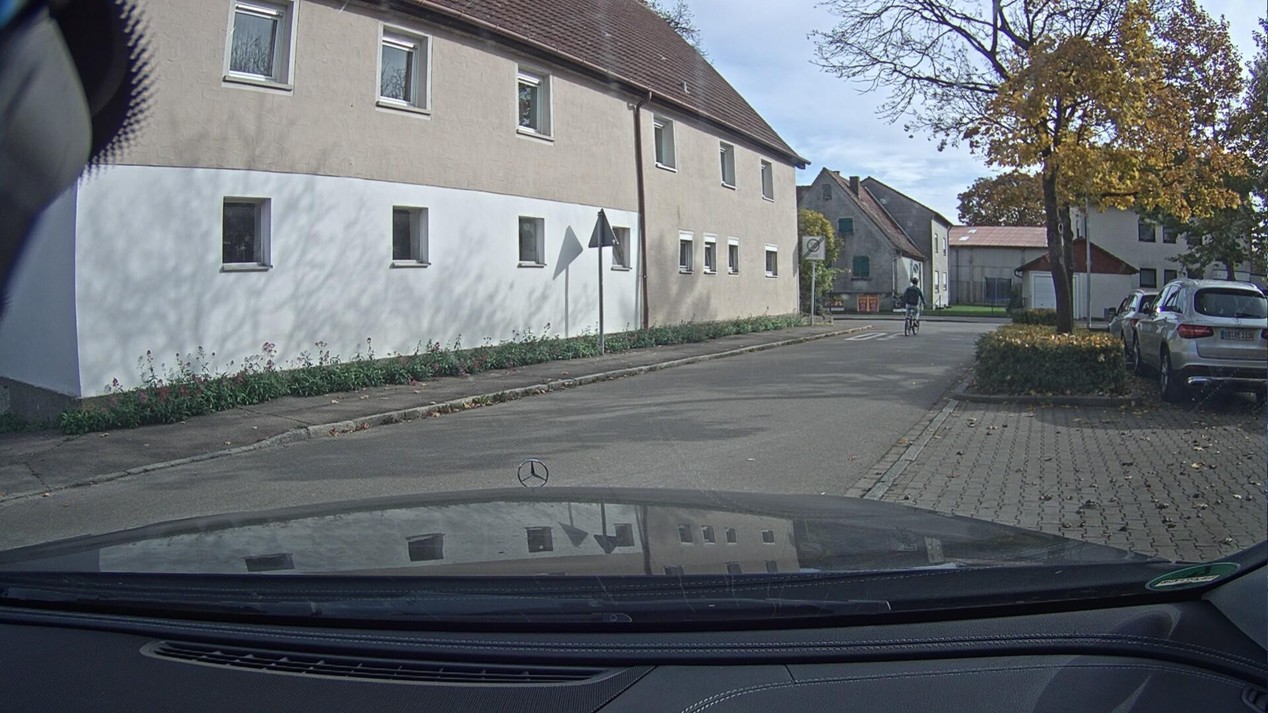}
		\captionsetup{skip=1pt} \caption*{House Facade}
	\end{subfigure}

	\begin{subfigure}{.31\columnwidth}
		\includegraphics[width=\columnwidth]{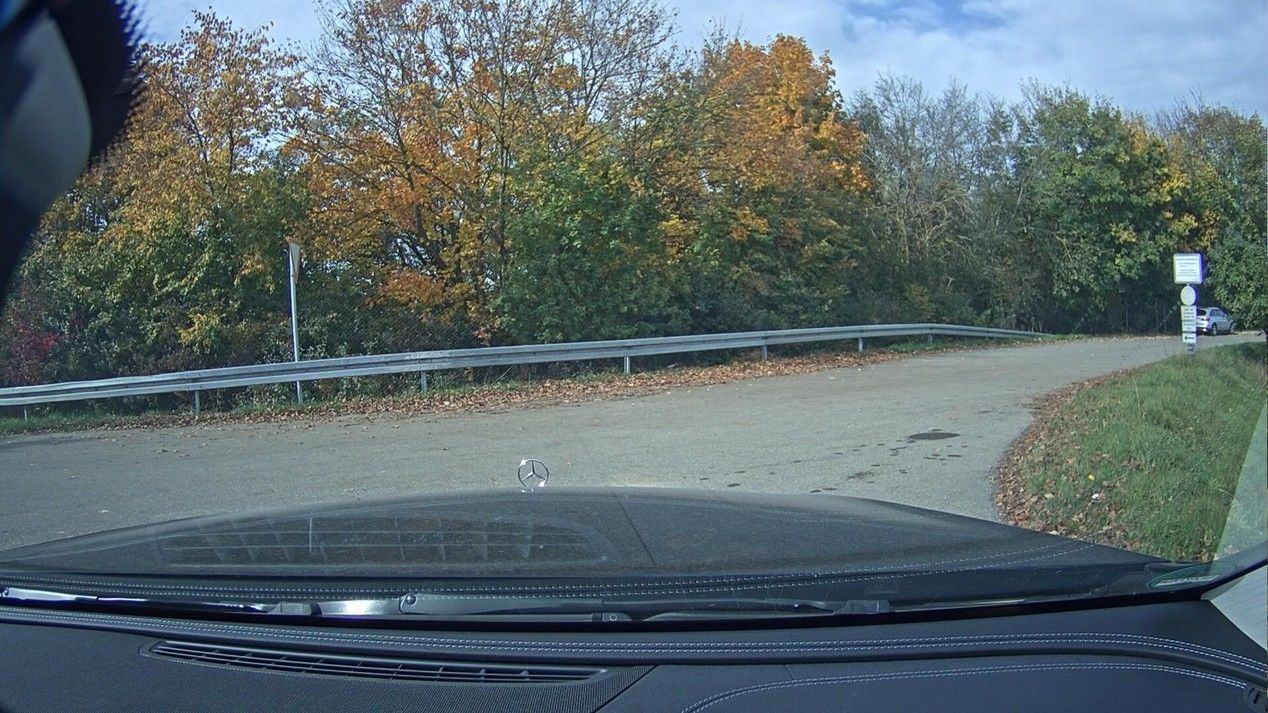}
		\captionsetup{skip=1pt} \caption*{Guard Rails}
	\end{subfigure}
	\begin{subfigure}{.31\columnwidth}
		\includegraphics[width=\columnwidth]{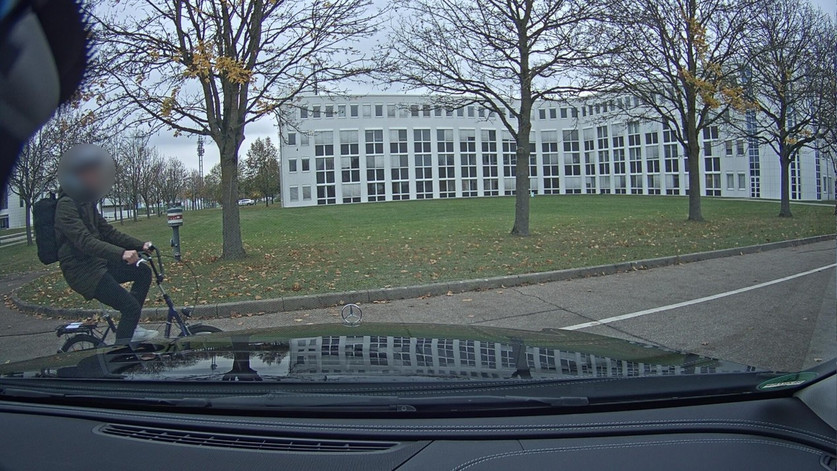}
		\captionsetup{skip=1pt} \caption*{Curbstone}
	\end{subfigure}
	\begin{subfigure}{.31\columnwidth}
		\includegraphics[width=\columnwidth]{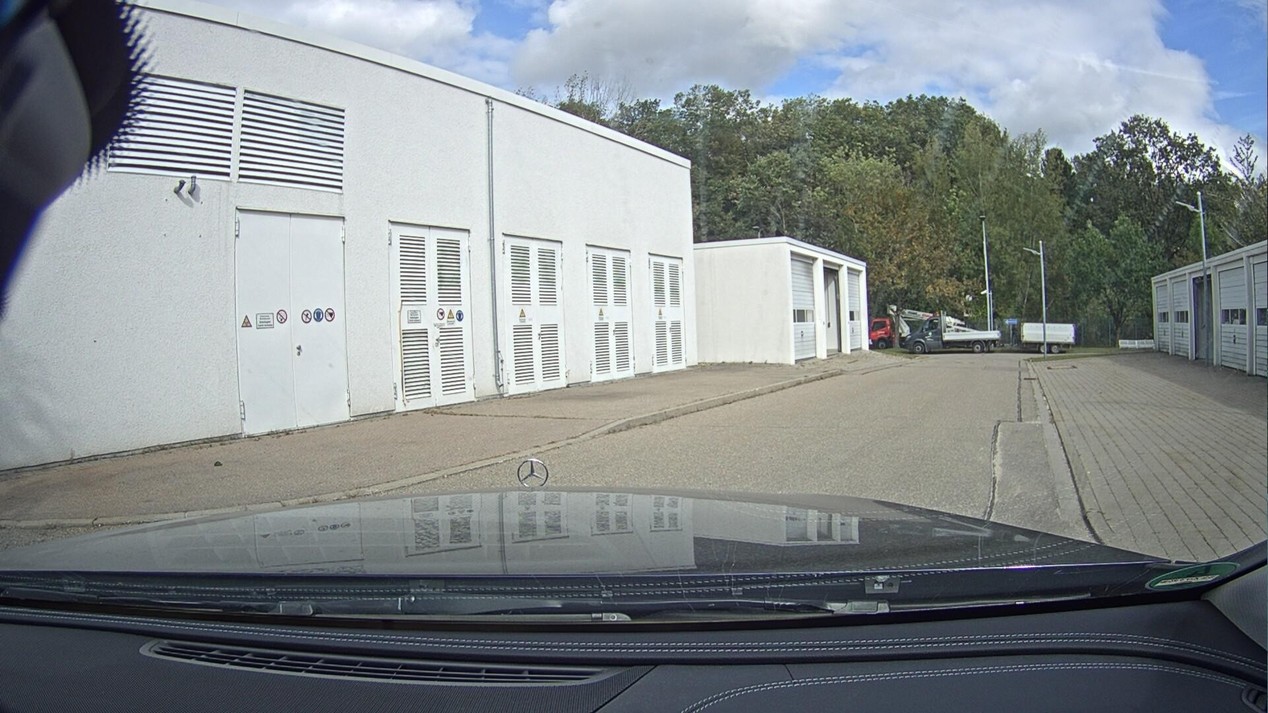}
		\captionsetup{skip=1pt} \caption*{Metal Doors}
	\end{subfigure}

	\caption{Recorded scenario and reflective surface examples.}
	\label{fig:dataset-impressions}
	\vspace{-5mm}
\end{figure}
 % label: fig:dataset-impressions
\section{Related Work}
\subsection{Radar Ghost Objects}
Ghost or multi-path reflections in radar data are a well known phenomenon \cite{Kamann2018, prophet2019RadarConf, Garcia2019ICMIM}.
Some ghost objects can be removed directly in the sensor \cite{Vermesan2013}, requiring access to the sensor and reducing the frame rate.
By using knowledge about the surrounding geometry ghost objects can be detected  \cite{sume2011radar,Qiu2014,zetik2015looking,rabaste2017around}.
Others propose a model based approach using a rigid body assumption \cite{Roos2017}.
In \cite{Scheiner2020CVPR}, ghost objects are used for a radar based non-line-of-sight application.
Machine learning approaches to detect and suppress ghost objects are presented by \cite{prophet2019RadarConf, Garcia2019ICMIM}.
In our previous work \cite{Kraus2020ITSC} we showed that it is possible to detect ghost objects using a semantic segmentation network.

\subsection{Radar Datasets}
In addition to our previously published NLOS dataset \cite{Scheiner2020CVPR}, there have been a couple of recent publications featuring automotive Doppler radar datasets. These datasets comprise Astyx \cite{astyxDataset}, nuScenes \cite{nuscenes2019}, CARRADA \cite{Carrada2020}, and Zendar \cite{zendar2020}.
Despite the increasing number, with the exception of \cite{Scheiner2020CVPR}, neither of those datasets include annotations for ghost objects.
As stated in the introduction, \cite{Scheiner2020CVPR} is focused on reconstructing occluded objects from the ghost images.
Therefore, a lidar system was used with the assumption that all radar points are multi-path objects if they fall behind barriers which are identified by the laser scanner.
This approach not only requires an additional sensor for identification but, also, completely neglects the possibility for multi-path occurrences which do not appear behind an obstacle.
Our dataset improves on the annotations in \cite{Scheiner2020CVPR} by adding detailed semantic instance annotations for different kinds of multi-path reflections.
This allows for conducting in-depth experiments for ghost suppression or exploitation.

\subsection{Point Cloud Processing}
Unordered point clouds can be processed in various ways.
In traditional radar processing a clustering algorithm is exploited to extract point conglomerates which represent object proposals \cite{Scheiner2019ITSC}.
These clusters can be used as input to classical classification models \cite{Scheiner2018IV}.
Another approach is to render the point cloud into a birds-eye-view (BEV) grid map and use image processing techniques \cite{Schumann2020TIV, Garcia2019ICMIM}.

The first neural network architectures handling unordered point clouds directly were PointNet \cite{Charles_Su_Kaichun_Guibas_2017} and its successor PointNet++ \cite{Qi2017}.
Ever since, various other architectures were proposed which follow the same paradigm, i.e., end to end point cloud processing \cite{Simonovsky_Komodakis_2017, Xu_2018_ECCV, Atzmon_Maron_Lipman_2018, NEURIPS2018_f5f8590c, Wu_Qi_Fuxin_2019, Thomas_Qi_Deschaud_Marcotegui_Goulette_Guibas_2019}.
SGPN \cite{wang2018sgpn} introduces instance segmentation for point clouds by using an embedding space and a similarity matrix to group multiple points into instances.

PointNet++ led to work directly processing radar point clouds end to end with deep learning based approaches.
In \cite{Schumann2018}, it was used to semantically segment radar data.
A recurrent instance segmentation architecture based on methods from PointNet++ was introduced \cite{Schumann2020TIV}.
Instances are build by means of an instance classifier module which is trained in a secondary training phase, utilizing a predicted per-point center offset.
In \cite{Danzer2019ITSC} region of interest proposals are used which are then processed by PointNet architectures to regress a 2D bounding box.

\begin{figure}[tb]
	\renewcommand{\arraystretch}{1.3}
	\centering

	\begin{subfigure}{.52\columnwidth}
		\resizebox*{\columnwidth}{!}{
			\begin{tabular}[b]{@{}cccc@{}}
				\toprule
				$f / \SI{}{\giga\hertz}$ & $r / \SI{}{\meter}$ & $\yaw / \deg$ & $\radvel /\SI{}{\meter\per\second}$\\
				$77$ & $0.15-153$ & $\pm70$ & $\pm 44.3$ \\
				\midrule
				
				$\Delta_t / \SI{}{\milli\second}$ & $\Delta_r / \SI{}{\meter}$ & $\Delta_\yaw /\deg$ & $\Delta_{\radvel} /\SI{}{\meter\per\second}$\\
				$100$ & $0.15$ & $1.8$ & $0.087$\\
				\bottomrule	
			\end{tabular}
		}
	\end{subfigure}
	\begin{subfigure}{.44\linewidth}
		\includegraphics[width=\columnwidth]{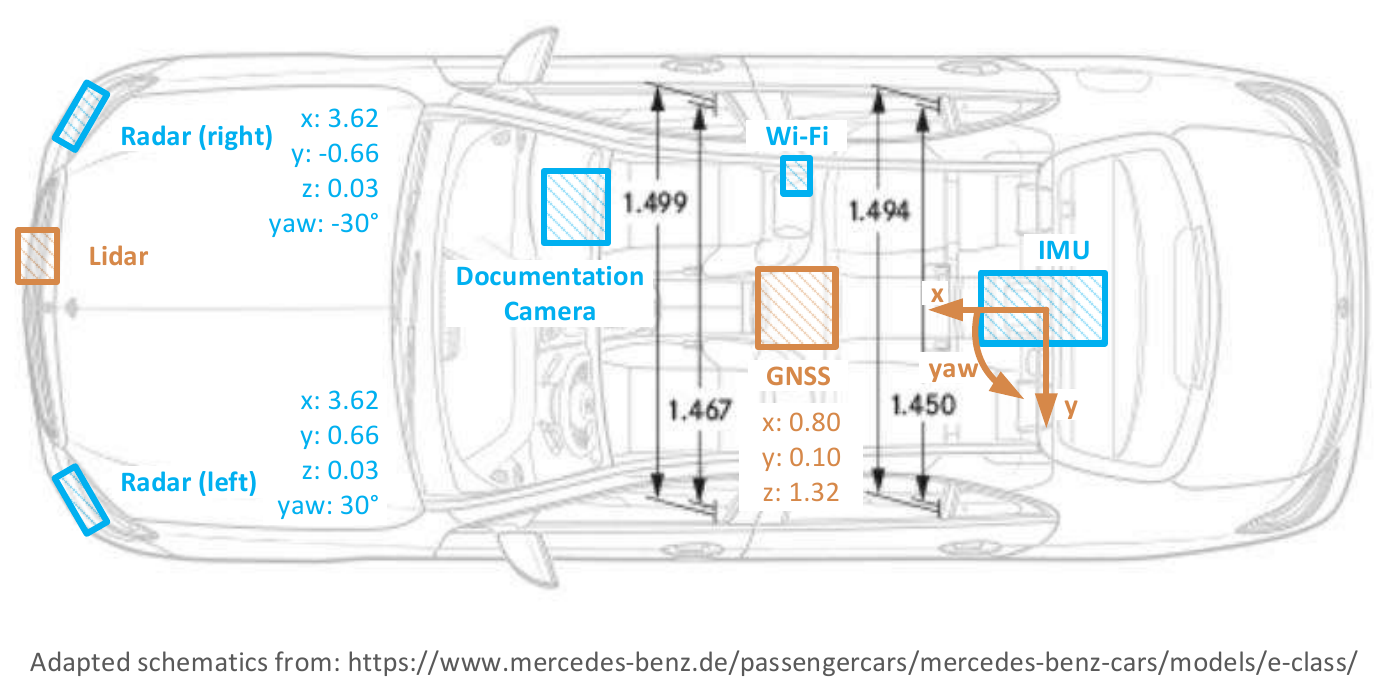}
	\end{subfigure}

	\caption{
		Radar sensor specification and sensor mounting positions schematic.
		Upper: Frequency band $f$, range $r$, azimuth angle $\yaw$, and radial (Doppler) velocity $\radvel$.
		Lower: Resolutions  $\Delta$ for $r$, $\yaw$, $\radvel$, and time $t$.
		Adapted from \cite{Scheiner2020CVPR} and \cite{Kraus2020ITSC}.
	}
	\label{tab:sensor-specs}
    \vspace{-6mm}
\end{figure}
 % \label{tab:sensor-specs}

\section{Radar Ghost Objects}
\label{sec:radar-ghosts}
Chirp sequence radars send out multiple frequency ramps which are reflected by surrounding objects.
The backscattered waves are picked up by a receiver antenna array and arranged according to the received time and antenna location within the array.
After undergoing several frequency transformations a 3D data cube is returned.
The different bins of this data cube resolve range, angle and Doppler velocity with the value corresponding to the received amplitude.
The Doppler velocity is the radial velocity of an object relative to the sensor.
An in depth analysis can be found in \cite{Scheiner2020CVPR}.

Due to the size and the fact that most entries in this data cube contain background noise the data cube is filtered with a constant false alarm rate (CFAR) filter \cite{Rohling1983}.
CFAR is an adaptive thresholding algorithm to suppress detections with low amplitude relative to their neighborhood.
Typically CFAR is applied in all dimensions of the 3D data cube independently, starting with range and angle.
As a result Doppler bins corresponding to non-stationary objects are more likely to surpass the CFAR detector.
This results in a good separation between moving and stationary objects.

Ghost detections are the result of multi-path propagations from the radar signal.
They occur whenever a multi-path reflection is received by the sensor and simultaneously surpasses the adaptive CFAR filtering.
This is most common when the multi-path reflection is reflected by a surface which is flat relative to the wavelength of the signal.
Due to the large wavelength of around \SI{4}{\mm} for a typical \SI{77}{\giga\hertz} radar, many surfaces act as mirrors by preserving a specular component of the signal.

As mentioned earlier, non-stationary objects have a higher chance of surpassing the CFAR filter.
This is also valid for the preservation of multi-path reflections caused by moving objects.
Consequently, ghost objects caused by moving objects are more common.
This fact was exploited for a non-line of sight application with automotive radar in \cite{Scheiner2020CVPR}.

As in \cite{Kraus2020ITSC} we distinguish between two fundamental types of multi-path reflections:
Those reflections where the last bounce happens on the object and those where it bounces off the reflective surface.
To distinguish between both, we use the convention from \cite{Liu2016b} and refer to them as type-1 and type-2 reflections.
Type-1 detections bounce back from the object and type-2 from the reflective surface.
Furthermore, multi-path reflections are categorized by the number of objects they bounce off.
We call this the \textit{order of the bounce}.
A signal which takes a direct round trip from the sensor to the object and back to the sensor is called a real or first order reflection.
Each additional bounce increases the order resulting in second and third order bounces
A real world example and schematic are shown in \autoref{fig:title}.
It is possible for higher order bounces to surpass the CFAR threshold, however, this leads to the possibility of more than one intermediate surface reflecting the signal.
Strictly speaking, even a third bounce could be reflected by different surfaces on both ways.

In this work, we focus on ghost detections caused by a single reflective surface.
Those have the highest chance to be preserved.
Furthermore, type-2 third bounce reflections can occur even if no direct line of sight between the sensor and the object exists.
Each time the signal bounces off a surface it is split into a diffuse and a specular reflection part.
Specular reflections emerge with the same angle as the incident signal relative to the surface normal.
All other reflections are called \textit{diffuse}.
Diffuse scattering comes with a wider spreading of the radar's energy.
Since not all energy is preserved in the signal, the amount of detected higher order reflections is decreased.
In this context, specular reflections are considered relative to the wavelength of the sensor (around \SI{4}{\mm}).

\begin{figure*}[ht]
	\centering
	\begin{subfigure}{.495\textwidth}
		\includegraphics[width=\textwidth]{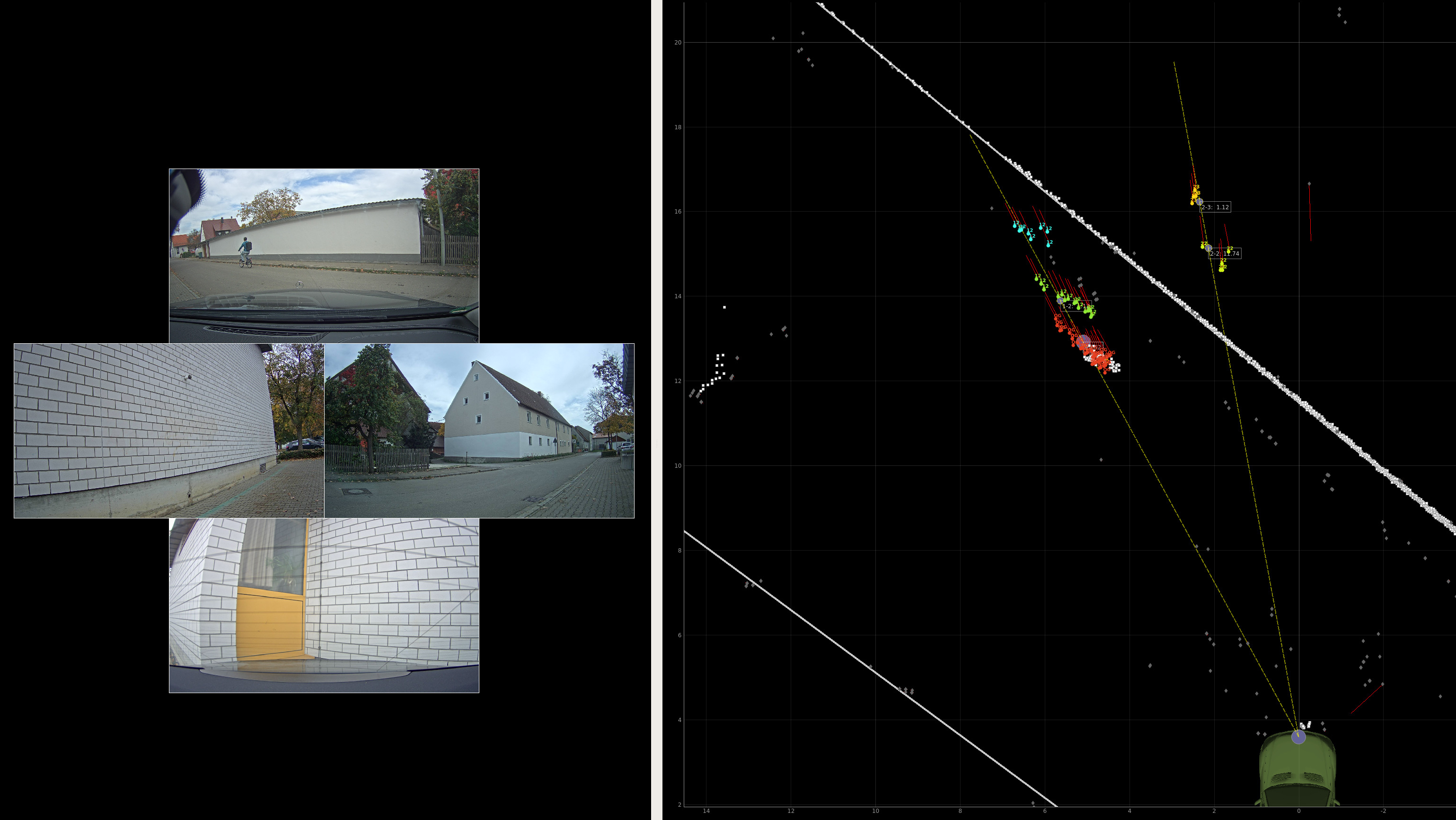}
		\caption{Screenshot of the labeling tool.
			On the left, images of the documentation cameras.
			On the right, the corresponding radar and lidar data.}
		\label{subfig:labeltool}
          \vspace{2mm}
	\end{subfigure}
	\begin{subfigure}{.495\textwidth}
	\includegraphics[width=\textwidth]{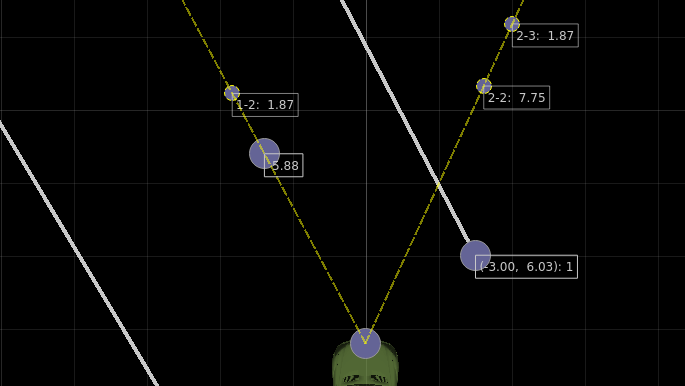}
	\caption{A multi-path helper tool indicates potential multi-path reflection locations based on the positions of the sensor, the real object and the reflective surface.}
	\label{subfig:multi-path-helper}
        \vspace{2mm}
    \end{subfigure}
	\begin{subfigure}{\textwidth}
		\begin{subfigure}{.195\textwidth}
			\includegraphics[width=\textwidth]{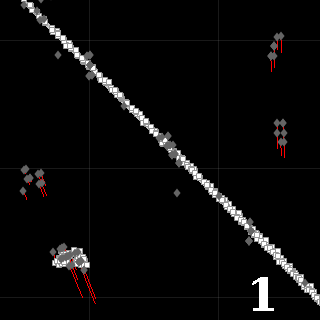}
		\end{subfigure}
		\begin{subfigure}{.195\textwidth}
			\includegraphics[width=\textwidth]{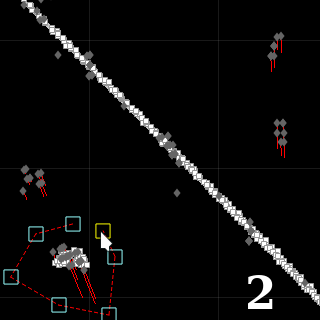}
		\end{subfigure}
		\begin{subfigure}{.195\textwidth}
			\includegraphics[width=\textwidth]{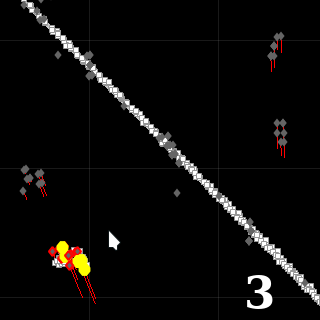}
		\end{subfigure}
		\begin{subfigure}{.195\textwidth}
			\includegraphics[width=\textwidth]{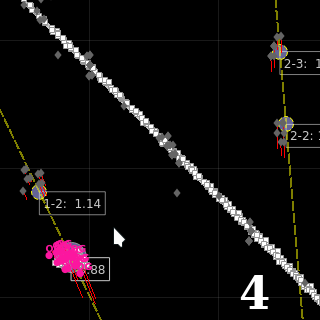}
		\end{subfigure}
		\begin{subfigure}{.195\textwidth}
			\includegraphics[width=\textwidth]{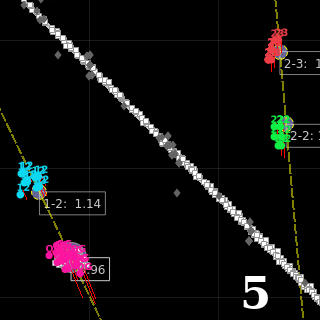}
		\end{subfigure}
	\caption{
		Example of labeling a frame with multi-path reflections: 1) shows the unlabeled frame and 2) the object selection process. In 3), the object is selected and assigned a label, which becomes visible in 4). Now, the multi-path helper snaps into place. After assigning the corresponding labels, the final labeled scene is indicated in 5).
	}
	\label{sugfig:annotation-process}
	\vspace{2mm}
	\end{subfigure}
	\begin{subfigure}{.329\textwidth}
		\includegraphics[width=\textwidth]{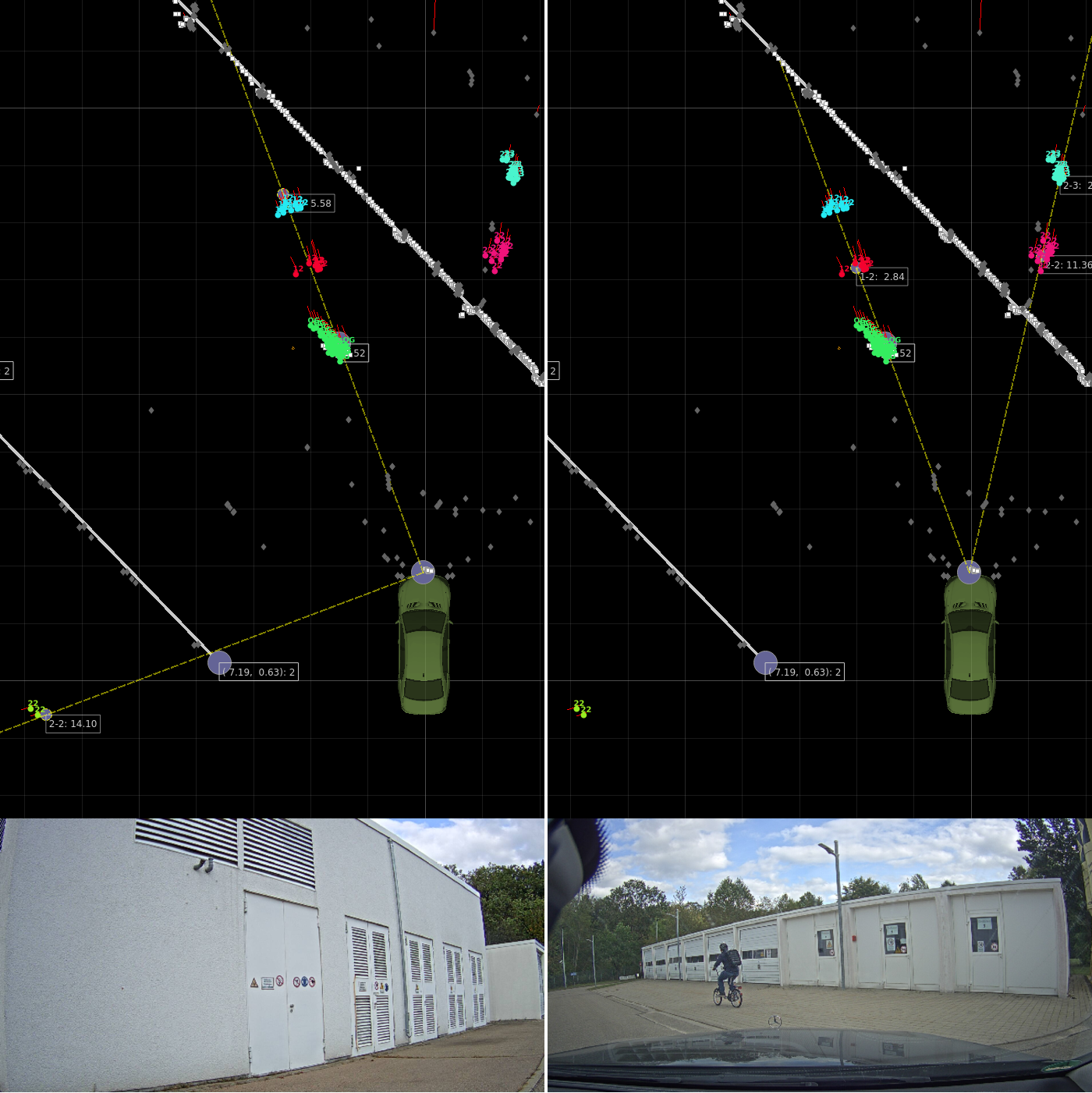}
			\caption{Two type-1 second order bounces caused by two walls.
			The multi-path helper indicates which reflections correspond to which reflective surface.}
		\label{fig:multi-wall}
	\end{subfigure}
	\begin{subfigure}{.329\textwidth}
		\includegraphics[width=\textwidth]{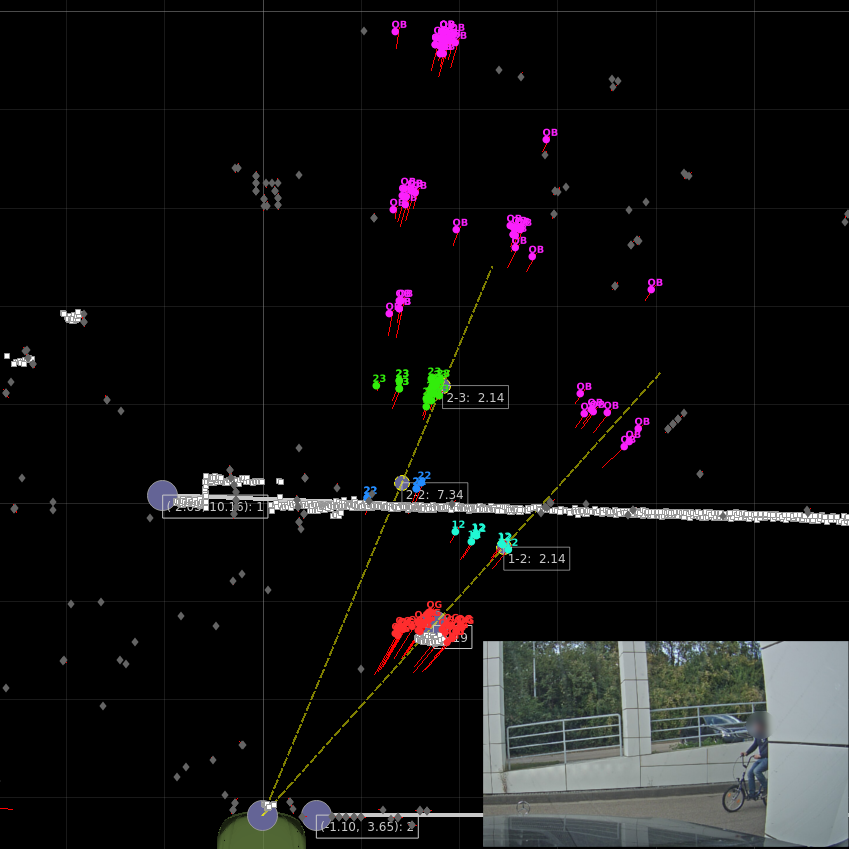}
		\caption{Scene with higher order multi-path reflections (pink), caused by two opposing highly reflective walls.}
		\label{subfig:other-bounces}
	\end{subfigure}
	\begin{subfigure}{.329\textwidth}
		\includegraphics[width=\textwidth]{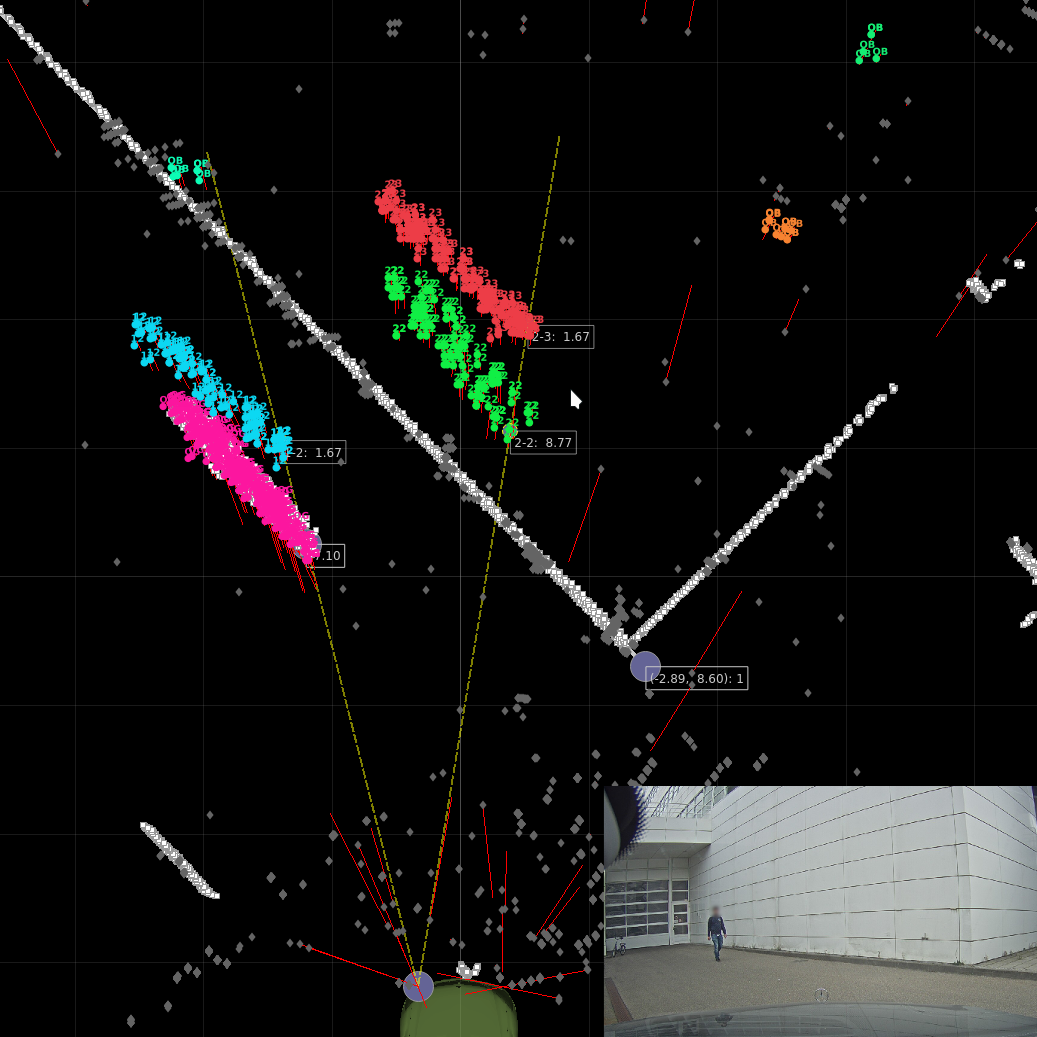}
		\caption{Accumulation over 20 frames: Four main objects emerge (pink: real, blue: mp-12, green: mp-22, red: mp-23).
			OMP reflections in the background.
			}
		\label{subfig:label-example-accumulated}
	\end{subfigure}
	\caption{Overview of the labeling process. The red lines on the radar detections are a visualization of the Doppler velocity, longer lines correspond to higher Doppler values. Cf. \autoref{tab:annotations} for abbreviations. Best viewed on screen.}
	\label{fig:labeltool-overview}
	\vspace{-4mm}
\end{figure*}
 % multiple labels

\begin{table}[tb]
	\caption{General statistics for the 111 annotated sequences.}
	\label{tab:seq-statistics}
	\centering
	\begin{tabular}{@{}rcccc@{}}
		                                               & \textbf{min} & \textbf{max} & \textbf{mean} & \textbf{median} \\ \midrule
		\textbf{Frames per Sequence} (\SI{10}{\hertz}) &    $163$     &    $716$     &   $385.52$    &      $371$      \\
		               \textbf{Radar Points per Frame} &    $127$     &    $1775$    &   $819.91$    &      $765$      \\
		                \textbf{Annotations per Frame} &     $0$      &    $156$     &    $22.59$    &      $17$
	\end{tabular}
\vspace{-5mm}
\end{table} % \label{tab:seq-statistics}
\section{Dataset}
\label{sec:dataset}
Our dataset is based on the data recorded for \cite{Scheiner2020CVPR}.
The original dataset has annotations created by an automatic labeling system \cite{Scheiner2019IRS}.
For this article, we provide manual and detailed annotations for all types of multi-path reflections.
Moreover, we publish additional sequences not included in the original dataset, plus additional synthesized sequences\footnote{Dataset available via \url{https://github.com/flkraus/ghosts}}.

In total, the dataset consists of 21 different scenarios. Each scenario is repeated between two and eight times.
From this we manually annotated 111 sequences.
Each recording contains a main VRU (vulnerable road user) object.
This is either a pedestrian or cyclist.
Having one main object per sequence makes it possible to accurately associate multi-path reflections with a real object.
For each sequence the main object moves away from the ego-vehicle until out of sight and then returns.
The path taken by the main object is near one or multiple reflective surfaces.
To prolong the time where multi-path reflections occur, the ego vehicle remains stationary during recordings.

The reflective surfaces in our recordings are: building and garden walls from different materials (plastered, marble, metal containers, mixed), parked cars, a concrete curbstone, and a guard rail.
Example images from the scenarios are displayed in \autoref{fig:dataset-impressions}.
A full overview of all 21 scenarios is given on the dataset homepage.

In total, the provided 111 annotated sequences from 21 scenarios amount to \SI{71}{\minute} of data.
On average, a single recording takes \SI{38.5}{\second} corresponding to 385 frames at \SI{10}{\hertz} sampling frequency of our radar setup, cf. \autoref{tab:sensor-specs} for the sensor specification and \autoref{tab:seq-statistics} for a more detailed dataset statistic.
The data was recorded using two experimental radar sensors mounted in the front bumper of our test vehicle.
Furthermore, four wide angle cameras and a four layer lidar sensor provide additional data, cf. \autoref{tab:sensor-specs}.
An in-depth description of the recording process can be found in \cite{Scheiner2020CVPR}.

The dataset contains around 35 million radar points and 1 million manual annotations.
Among those annotated points roughly 100 thousand correspond to ghost objects.
Another 600 thousand points correspond to real objects.
Statistics for the most important annotations are given in \autoref{tab:annotations}.

\vspace*{-0.85mm}
\subsection{Annotations}
The original dataset contains annotations for real objects and type-2 third order bounces.
A main contribution of this article is the publication of detailed semantic instance annotations for different kinds of multi-path reflections.
This extension of the original annotations is most important when the task at hand is to segregate radar multi-path detections without any other sensors such as the lidar used in \cite{Scheiner2020CVPR}.

For each sequence, the main object is annotated.
We differentiate between \emph{real detections}, \emph{type-1} or \emph{type-2 second order bounces}, and \emph{type-2 third order bounces}, visual example given in \autoref{subfig:label-example-accumulated}.
Furthermore, we annotate all higher order bounces caused by the main object as \emph{other multi-path bounces}, as depicted in \autoref{subfig:other-bounces}.
If multiple reflective surfaces are present in a scene, we annotate which surface causes a reflection (only for detailed bounces).
In some cases different bounce types are not distinguishable from one another.
For example, if the object is close to the wall, it is not always clear if the observed multi bounce is a second or third order bounce.
This is annotated as \emph{indistinguishable}.

During the recording process great care was taken that no other objects interfere with the recordings.
If in some cases additional objects, other than the main object, are visible in the scene they are also annotated as VRUs or motorized vehicles.
To not interfere too much with the annotation process multi-path reflections caused by those objects are annotated as \emph{other multi-path bounces} without further differentiation into different bounce types.

Each annotation also has a class tag.
VRU objects are either \emph{pedestrian} or \emph{cyclist} and vehicles \emph{car}, \emph{large vehicle}, and \emph{motorbike}.
For ambiguous annotations we add a \emph{sketchy} tag.
Lastly, for indeterminable regions or reflections we add an \emph{ignore} label.
All data points without an explicit label are considered background.
A statistic for the annotations and a shortened list of the most important labels and all class tags is provided in \autoref{tab:annotations}.

\begin{table}[tb]
	\caption{Number of annotated radar points. Only the main annotations are shown excluding annotations with the \emph{sketchy} tag. In parenthesis: Instances consisting of at least 3 points.
	}
	\label{tab:annotations}
\centering
\resizebox{\columnwidth}{!}{
	\begin{tabular}{@{}rcrrrrr@{}}
	\toprule
	 \textbf{Object} & \textbf{Class} &        \textbf{Real} & \textbf{OMP} &     \textbf{MP-12} &     \textbf{MP-22} &     \textbf{MP-23} \\ \midrule
	   \textbf{Main} &      ped       & $316,990$ ($17,502$) &      $6,748$ & $23,236$ ($3,157$) & $20,008$ ($2,764$) & $22,735$ ($2,332$) \\
	                 &      cycl      & $253,597$ ($10,472$) &      $5,575$ & $14,011$ ($1,711$) & $12,586$ ($1,577$) & $17,195$ ($1,548$) \\ \midrule
	  \textbf{Other} &      ped       &   $41,775$ ($5,038$) &         $94$ &                    &                    &                    \\
	                 &      cycl      &         $202$ ($29$) &        $214$ &                    & \multicolumn{2}{c}{\textbf{background}} \\ \cmidrule(r){0-3} \cmidrule(l{2mm}r{2mm}){6-7}
	\textbf{Vehicle} &      car       &     $14,328$ ($740$) &      $1,073$ &                    &    \multicolumn{2}{c}{$34,073,595$}     \\
	                 &     truck      &       $2,218$ ($77$) &          $0$ &                    &                    &                    \\
	                 &      bike      &          $134$ ($5$) &         $18$ &                    &                    &                    \\ \bottomrule
\end{tabular}
}
	\vspace{2mm}
	\captionsetup{justification=centering,margin=.5cm}
	\caption*{\textbf{Classes}: \textbf{Ped}estrian, \textbf{Cycl}ist, \textbf{Car}, \textbf{Truck}, Motor\textbf{bike}. \textbf{Labels}: \textbf{MP} stands for \textbf{M}ulti-\textbf{P}ath the first digit represents the type and the second the order. \textbf{OMP} stands for \textbf{O}ther \textbf{M}ulti-\textbf{P}ath, i.e., unspecific multi-path detections.}
\vspace{-5mm}
\end{table} % \label{tab:annotations}

\vspace*{-0.85mm}
\subsection{Annotation Process}

For the annotation process a labeling tool is used which shows a birds-eye-view (BEV) of the radar point cloud, cf. \autoref{subfig:labeltool}.
The radar data is overlaid with additional lidar data and the position of the GNSS reference system.
The images from the documentation cameras are shown alongside the other data.
Since pure radar data is difficult to annotate, having additional data from the other sensors available is crucial.

To make the annotation process possible, we implement a multi-path helper tool, cf. \autoref{subfig:multi-path-helper}.
This tool calculates the position of possible specular second and third order bounces
along a given wall.
The calculations are based on the position of the real object, the sensor and the reflective surface.
The annotation helper is an overlay to the radar view and consists of interactive walls and a movable object.
Each wall can be adjusted via its end points.
The sensor position is fixed at the front of the ego-vehicle.
The helper tool allows to switch between multiple reflective surfaces.
This is useful if different walls simultaneously cause multi-path reflections, e.g., as in \autoref{fig:multi-wall}.
In scenarios with multiple reflective surfaces, we annotate which wall causes a bounce.
To improve the labeling process, the multi-path helper automatically adjusts itself once the real reflections of the main object are annotated.

Semantic instances are annotated, i.e., for each object instance all radar points are grouped together and assigned a label.
Whenever possible the same instance is kept over multiple frames as show in  \autoref{subfig:label-example-accumulated}.
An object instance is created by selecting an area in the BEV and assigning a label.
To speed-up the annotation process, several time frames can be accumulated and processed at the same time.
The labeling process is shown in \autoref{sugfig:annotation-process}.

\subsection{Synthesized Dataset}
Due to the extreme effort for recording and annotation, our dataset only contains a single main object for each sequence, ignoring the occasional appearance of other objects.
To alleviate this shortcoming, we utilize a simple yet effective synthesizing technique for overlaying multiple objects.
Since our ego vehicle is stationary, it is possible to overlay multiple recordings from a single scenario with each other.
It is even possible to overlay a recording with itself by starting at different times within the recording.

\begin{figure}[tb]
	\centering
	\includegraphics[width=\columnwidth]{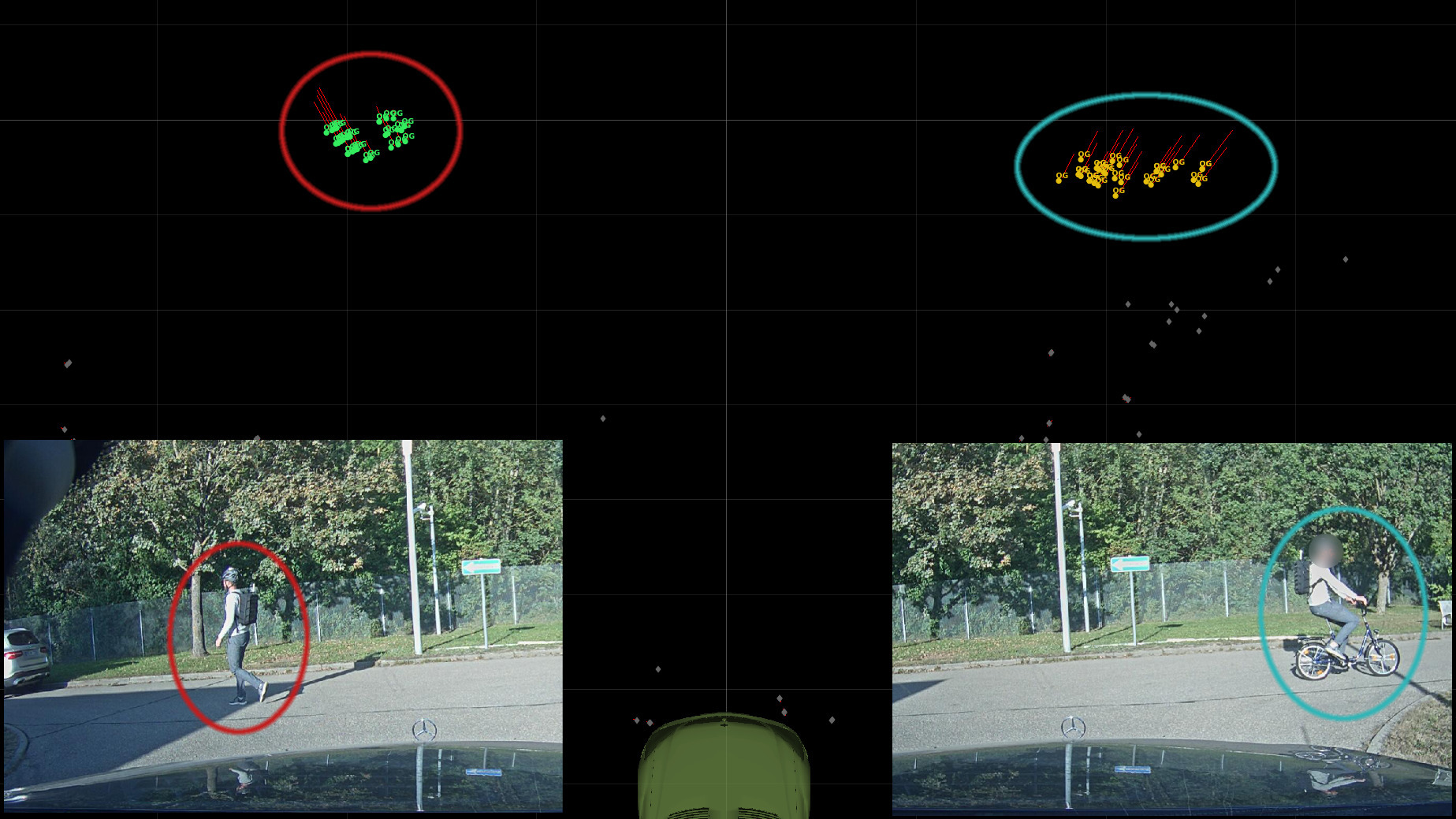}
	\caption{Two different sequences of the same scenario overlaid.}
	\label{fig:overlay-example}
	\vspace{-5mm}
\end{figure}

During the overlay process, great care is taken to avoid potential overlaps between annotated objects.
The synthesized sequences consist all of at least ten frames, between two and five different sequences are overlaid.
An example frame from an overlaid sequence is shown in \autoref{fig:overlay-example}.

\begin{figure*}
	\centering
	\begin{minipage}{0.69\linewidth}
		\centering
		\captionsetup{type=table} %% tell latex to change to table
		\caption{Results for \emph{pedestrian} vs. \emph{cyclist}. Average precision (AP) per class and average (mAP). Scores are given at two different IoU thresholds.}
		\label{tab:instance-seg-res-ped-cycl}
		\resizebox{\linewidth}{!}{
			\begin{tabular}{@{}rcccccccccccc@{}}
				\toprule
				    \textbf{Score} & \textbf{Method} &  \textbf{Ped}  & \textbf{Ped-12} & \textbf{Ped-22} & \textbf{Ped-23} & \textbf{Ped-Ghost} & \textbf{Cycl}  & \textbf{Cycl-12} & \textbf{Cycl-22} & \textbf{Cycl-23} & \textbf{Cycl-Ghost} & \textbf{Average} \\ \midrule
				      \textbf{0.3} &  \textbf{SGPN}  &     79.34      &                 &                 &                 &                 &     77.58      &                  &                  &                  &                  &      78.46       \\
				                   &                 &     70.89      & \textbf{65.98}  & \textbf{41.28}  &      31.70      &                 &     70.43      &  \textbf{50.93}  &  \textbf{52.73}  &  \textbf{32.71}  &                  &       52.0       \\
				                   &                 &     71.47      &                 &                 &                 &      45.72      &     77.22      &                  &                  &                  &      45.18       &      59.90       \\
				\cmidrule(r){2-13} & \textbf{DBSCAN} & \textbf{83.79} &                 &                 &                 &                 & \textbf{79.45} &                  &                  &                  &                  &      81.62       \\
				                   &                 &     78.90      &      51.10      &      34.38      & \textbf{34.91}  &                 &     72.85      &      39.49       &      39.34       &      29.93       &                  &      47.60       \\
				                   &                 &     79.90      &                 &                 &                 & \textbf{56.13}  &     74.83      &                  &                  &                  &  \textbf{46.60}  &      64.17       \\ \midrule
				      \textbf{0.5} &  \textbf{SGPN}  &     78.49      &                 &                 &                 &                 &     75.87      &                  &                  &                  &                  &      77.18       \\
				                   &                 &     70.50      & \textbf{58.86}  & \textbf{35.30}  &      29.56      &                 &      69.0      &  \textbf{48.30}  &  \textbf{48.25}  &  \textbf{30.30}  &                  &      48.73       \\
				                   &                 &     71.10      &                 &                 &                 &      44.82      &     69.30      &                  &                  &                  &  \textbf{41.41}  &      56.66       \\
				\cmidrule(r){2-13} & \textbf{DBSCAN} & \textbf{83.64} &                 &                 &                 &                 & \textbf{78.82} &                  &                  &                  &                  &      81.23       \\
				                   &                 &     78.76      &      50.98      &      34.32      & \textbf{34.82}  &                 &     72.44      &      38.88       &      33.56       &      29.56       &                  &      46.66       \\
				                   &                 &     78.85      &                 &                 &                 & \textbf{56.20}  &     74.42      &                  &                  &                  &      39.80       &      62.27       \\ \bottomrule
			\end{tabular}
		}
	\end{minipage}
	\begin{minipage}{0.29\linewidth}
		\includegraphics[width=\linewidth]{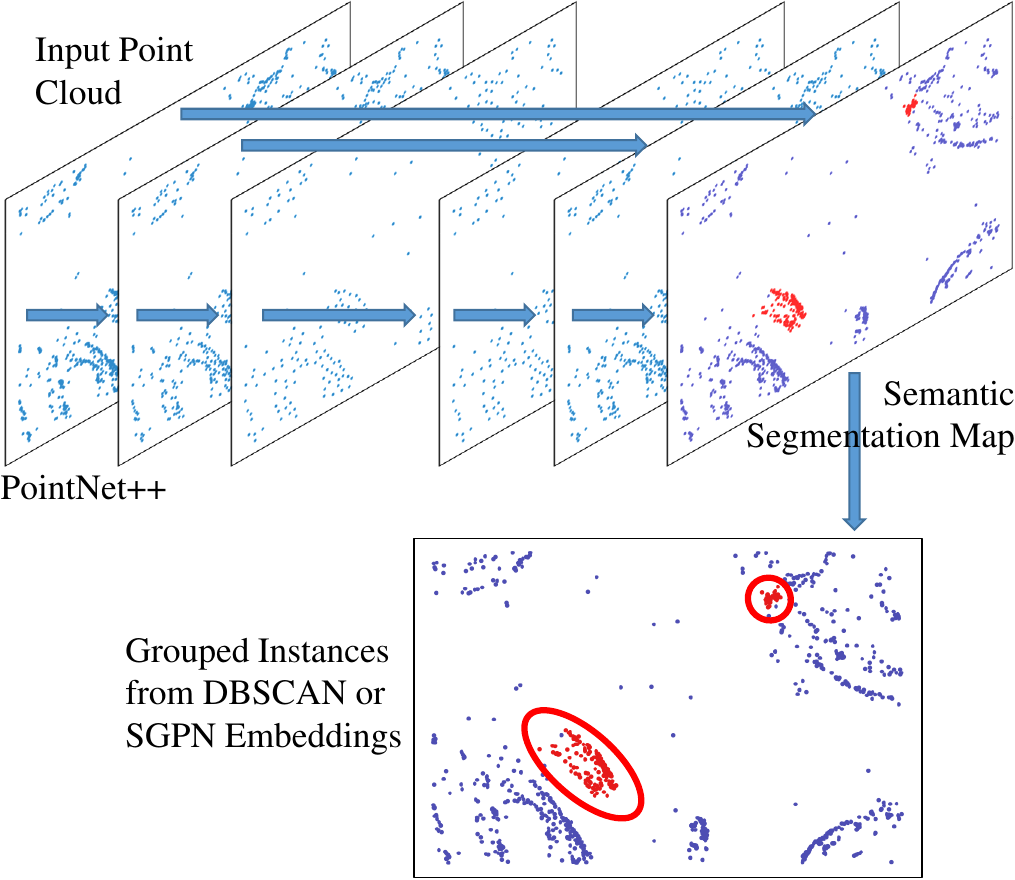}
		\caption{Object detection schematic.}
		\label{fig:pnpp-schematic}
	\end{minipage}
\vspace{-5mm}
\end{figure*}
 % \label{tab:instance-seg-res-ped-cycl}

\section{Methods}
\subsection{Semantic Segmentation with PointNet++}
\label{subsec:pnpp}
For our base segmentation network we utilize a PointNet++ \cite{Qi2017} variant optimized for radar data.
PointNet++ is a neural network architecture able to directly process point clouds in an end to end fashion.
For our setup we use three \emph{multi scale grouping} (MSG) modules for downsampling, followed by three \emph{feature propagation} (FP) modules for upsampling to the original point cloud size.
After the FP modules, the features are processed by \emph{fully connected} (FC) layers with a final softmax layer to classify each point.
This architecture resembles an hour glass structure used for image based semantic segmentation.
More details and specific parametrization can be found in \cite{Kraus2020ITSC, Schumann2018} and the original work \cite{Qi2017}. A schematic is given in \autoref{fig:pnpp-schematic}.

The input point cloud size is fixed to 2560, with points being accumulated over three measurement cycles (frames).
To achieve a fixed point cloud size the original point cloud is either up- or downsampled.
For upsampling points with the largest Doppler values are duplicated.
For downsampling points with low Doppler values are removed, with those from older measurement cycles being discarded first.

To account for class imbalances in the dataset, the classification softmax loss for each point is scaled relative to the inverse proportion each class takes up in the dataset.
Resulting in a scaling factor
\begin{equation}
s = \frac{1}{(c \cdot s_l)},
\label{eq:freq-balancing}
\end{equation}
with the total number of classes $c$ and the proportion $s_l$ of the occurrence of class $l$ in the dataset.

 % \label{fig:overlay-example}
\subsection{Instance Segmentation with SGPN}
\label{subsec:sgpn}
The Similarity Group Proposal Network (SGPN) is a method for point cloud instance segmentation.
It takes the output of a semantic segmentation network and splits it into three branches.
One is the original semantic class prediction, the second an embedding (or hash) prediction, and the third and final an object (or group) confidence.
In case of a PointNet++ base architecture, the different heads are attached right before the FC layers, keeping the original semantic segmentation branch.

For a point cloud size of $n \in \N$ and $c \in \N$ classes, the different network outputs are as follows.
An $n \times c$ class prediction, an $n \times 128$ embedding and an $n \times 1$ confidence score.
From the embeddings an $n \times n$ similarity matrix $S$ is calculated consisting of the pairwise distances between the embeddings for each point pair.

From $S$, the actual loss is calculated.
If two points are part of the same object instance the embedding difference is optimized towards $0$.
To this end two hinge losses with two thresholds $K_2 > K_1$ are utilized.
Where points of different semantic classes and instances need to be further away ($K_2$) in the embedding space then points in different instances but the same semantic class ($K_1$).
As a result, the $i$-th row of the similarity matrix $S$ represents a proposal containing all points $p_j$ with $\norm{p_i - p_j} < K_1$.

If a point $p_i$ is part of an object, the corresponding row of $S$ should reflect this with a low distance to all other points $p_j$ also contained in this object.
The pointwise confidence score is then trained to estimate the overlap of the real object with the object predicted by each row of $S$.
With background points being regressed to predict a confidence value of $0$.

For our work, we adapt SGPN to radar object detection.
SGPN was developed for much denser and less noisy data which is not comparable to automotive radar data.
Since the general idea for SGPN is still valid, we redesign the network, to apply it for the task at hand.

In the original implementation background points receive no loss for the embedding and only for the confidence score.
In radar based object detection most points are background or noise which we do not want to interfere with the foreground.
Hence, we change the loss implementation such that the ($i$, $j$) entry in $S$ must be larger than $K_2$ if exactly one of $p_i$ or $p_j$ is a background point.
Simultaneously, we do not apply a loss for entries  ($i$, $j$) in $S$ where both $p_i$ and $p_j$ are background points.
Therefore, the network is free to assign similar embeddings to background points as long as they are different from all foreground points.

Since our dataset, as well as most object detection datasets, contains ignore labels we apply the following rule.
If $i$ corresponds to a point $p_i$ which is labeled as \emph{ignore} or \emph{sketchy}, no loss is applied for the whole $i$-th row and column of $S$.
As well as not applying any loss for the $i$-th entry of the semantic segmentation part and confidence prediction.

The original implementation applies a per batch frequency balancing to the embedding loss similar to \cref{eq:freq-balancing}, lessening the effect of different sized objects.
However, as we also apply a loss for background points, the dominating factor is not the different sizes of objects but the huge imbalance between background (bg) and foreground (fg).
With roughly \SI{98}{\percent} of points being background.
Therefore, we scale the loss for each entry ($i$, $j$) in $S$ depending on whether it is a (bg, fg) or a (fg, fg) pair.
This is implemented by counting the occurrences of both pair types per batch and scaling the loss based on those numbers similar to \cref{eq:freq-balancing}.

Similar to other object detection frameworks, we apply non-maximum suppression (NMS) to discard proposals that have a pointwise intersection over union (IoU) greater than $0.5$ with a proposal with higher confidence.
This differs from the original implementation where intersecting instances are merged rather than discarded, which makes more sense in a setting with less background points.
Moreover, we discard all proposals containing less than three points.
The predicted class and score are calculated as follows.
For each point of an instance, we determine the non-background class with the highest score.
The class with the highest number of occurrences is set as the predicted class.
Then the score for this winning class is averaged over all points and multiplied with the mean confidence score.

As a backbone for the SGPN detection head, we use the PointNet++ architecture from \autoref{subsec:pnpp}.
A schematic of both object detection approaches can be found in \autoref{fig:pnpp-schematic}.

\begin{table}
	\caption{Semantic segmentation F1 scores. Pedestrian and cyclist combined into object class. Different rows show models trained on different classes.}
\label{tab:semseg-res}
	\centering
	\begin{tabular}{cccccc}
		\toprule
		\textbf{Obj} & \textbf{Obj-12} & \textbf{Obj-22} & \textbf{Obj-23} & \textbf{Obj-Ghost} & \textbf{Average} \\ \midrule
		  $95.39$    &                 &                 &                 &                    &     $95.39$      \\
		  $93.78$    &     $71.56$     &     $60.99$     &     $52.25$     &                    &     $69.65$      \\
		  $91.38$    &                 &                 &                 &      $72.84$       &     $82.11$      \\ \bottomrule
	\end{tabular}
\vspace{-5mm}
\end{table}
 % \label{tab:semseg-res}

\section{Experiments and Results}
Preserving the splits presented in \cite{Scheiner2020CVPR}, the extended dataset is divided into train, val, and test splits containing 75, 8, and 28 sequences each.
The train and val split contain independent sequences from the same scenarios, whereas the test split has no scenario overlap with the train and val splits.

For all experiments we train six different models with different training classes.
Three models use the \emph{pedestrian} and \emph{cyclist} classes while the other three merge both classes into one \emph{object} class.
For both cases we train three different models.
First, a baseline model only separating real objects from background.
Second, a model which aims to predict real objects, and all three annotated multi-path types.
And lastly, a model which distinguishes between real and ghost objects, grouping the three multi-path types into a single \emph{ghost} class.

All models are trained on the original sequences as well as the synthesized data.
The synthesized data is necessary for the SGPN training since the embedding loss relies on the presence of multiple objects.
The evaluation for the instance segmentation is also performed on original and synthesized data.
Evaluating on the synthesized data with multiple objects ensures that SGPN is able to not only distinguish between foreground and background but between different instances as well.
The semantic segmentation results only use the original data.

\subsection{Semantic Segmentation}
First, we conduct semantic segmentation experiments similar to those in our earlier work \cite{Kraus2020ITSC} with our new annotations and a PointNet++ architecture described in \autoref{subsec:pnpp}.
We examine different input features and accumulation times.
Best results are obtained by accumulating the data over three sensor cycles (frames) and using the \emph{amplitude}, \emph{Doppler velocity}, and a \emph{relative timestamp} as features.
The relative timestamp is 0 for points belonging to the most recent sensor cycle and goes down in $-0.1$ decrements reflecting the \SI{10}{\hertz} sampling rate.

Results for the merged object class experiments using optimal settings can be found in \autoref{tab:semseg-res}.
Removing the Doppler velocity from the features, the F1 score for real objects drops to roughly $60\%$ while the type-2 classes only achieve a score of below $5\%$ and type-1 second bounce around $15\%$.
With most ghost detections being wrongly classified as background and some confused with real objects.
This dependence on the Doppler value confirms the findings in \cite{Schumann2018} and also indicates that ghost object detection depends much more on the Doppler feature than real objects.

\begin{table}
	\caption{Average precision (AP) per class and average (mAP). Pedestrian and cyclist combined into object class. Scores are given for two different IoU thresholds.}
\textbf{}	\label{tab:instance-seg-res-obj}
\centering
\resizebox{\columnwidth}{!}{
	\begin{tabular}{@{}rccccccc@{}}
		\toprule
		   \textbf{IoU} & \textbf{Method} &  \textbf{Obj}  & \textbf{Obj-12} & \textbf{Obj-22} & \textbf{Obj-23} & \textbf{Obj-Ghost} & \textbf{Average} \\ \midrule
		     \textbf{0.3} &  \textbf{SGPN}  & \textbf{86.99} &                 &                 &                 &                 &      86.99       \\
		                  &                 &     79.94      & \textbf{54.59}  & \textbf{50.78}  &      32.70      &                 &      54.35       \\
		                  &                 &     80.35      &                 &                 &                 & \textbf{57.46}  &      68.90       \\
		\cmidrule(r){2-8} & \textbf{DBSCAN} &     86.92      &                 &                 &                 &                 &      86.92       \\
		                  &                 &     78.69      &      49.33      &      44.60      & \textbf{43.24}  &                 &      53.83       \\
		                  &                 &     78.42      &                 &                 &                 &      54.99      &      66.71       \\ \midrule
		     \textbf{0.5} &  \textbf{SGPN}  &     83.86      &                 &                 &                 &                 &      83.86       \\
		                  &                 &     78.81      &      45.97      & \textbf{46.28}  &      30.85      &                 &      50.48       \\
		                  &                 &     79.56      &                 &                 &                 & \textbf{55.35}  &      67.46       \\
		\cmidrule(r){2-8} & \textbf{DBSCAN} & \textbf{86.90} &                 &                 &                 &                 &      86.90       \\
		                  &                 &     78.30      & \textbf{49.40}  &      43.21      & \textbf{42.84}  &                 &      53.35       \\
		                  &                 &     77.85      &                 &                 &                 &      54.48      &      66.16       \\ \bottomrule
	\end{tabular}
}
\vspace{-5mm}
\end{table}
  % \label{tab:instance-seg-res-obj}
\subsection{Instance Segmentation with DBSCAN and SGPN}
In addition to the end to end SGPN PointNet++ variant presented in \autoref{subsec:sgpn}, we extend our semantic segmentation network using a clustering algorithm as an alternative for instance segmentation.
We use an adapted DBSCAN variant \cite{Scheiner2019ITSC} optimized on our data as a second baseline.
The output of our semantic segmentation PointNet++ is used to filter the data.
If the network predicts a point as background with a confidence higher than $33\%$ it is removed.
The remaining points are then clustered by the DBSCAN variant.
Predicted clusters consist of at least three points.
For each cluster we calculate the prediction equal to our SGPN approach in \autoref{subsec:sgpn} by assuming a confidence score of $1.0$ for each point and only using the semantic class predictions.

The results of both instance segmentation approaches are presented in Tabs.~\ref{tab:instance-seg-res-ped-cycl} and \ref{tab:instance-seg-res-obj}.
We evaluate the instance segmentation experiments with the average precision (AP) metric \cite{PascalVOC}. % TODO cite
A predicted instance is considered as a true positive if it has an intersection over union (IoU) overlap greater than a given threshold.
The methods are evaluated at two different IoU thresholds 0.3 and 0.5.

SGPN performs better when it comes to segmenting different ghost object types which is most notable for the models detecting ghost objects and differentiating between the pedestrian and cyclist class.
Here we have a mean AP (mAP) over the ghost classes of $45.89\%$ ($41.76\%$) for the SGPN variant and $38.19\%$ ($37.02\%$) for the DBSCAN variant (values for IoU thresh 0.5 in parenthesis).
This difference equalizes to $46.02\%$ ($41.03\%$) versus $45.72\%$ ($45.15\%$) mAP for ghost objects with the combined \emph{object} class, mostly due to the fact that DBSCAN detects type2 third bounces much better than SGPN.

On the other hand DBSCAN outperforms SGPN on the real object class with a mAP for pedestrian and cyclists of $81.62\%$ ($81.23\%$) compared to  $78.46\%$ ($77.18\%$) for SGPN.
DBSCAN also seems to have an overall better localization, since the scores are more robust for the different IoU thresholds.

Regardless of the chosen method a respectable AP of around $40\%$ to $50\%$ is achieved on the challenging ghost object classes.
The hardest class to detect are type2 third order bounces with an AP of around $30\%$ for most models.
Lastly if we combine all different ghost types into a single ghost class an AP of around $55\%$ is achieved.
With the ghost cyclist class being the hardest with only around $45\%$.
This setting of combining all ghost objects into a single class is also the most likely to be annotated in larger scale radar datasets.
Since it is relatively easy to spot ghost objects but hard to correctly identify the order of the multipath reflection.

\subsection{Ghost Object Confusion}
Here we evaluate how the presence of ghost objects affects object detection algorithms trained without ghost annotations.
For this, we calculate the confidence score corresponding to the highest F1 score in the precision recall curve.
We use this score as a threshold for our detections.
The remaining detections are split into true positives (TPs) and false positives (FPs).
We then analyze what caused the FP detections.
The results are shown in \autoref{tab:confusion}.
It is notable that SGPN has less confusion due to ghost objects, however more FPs are caused by background detections or confusion of cyclists with pedestrians.
Overall, type1 second order bounces seem to cause the most problems with up to $36\%$ of FPs for the DBSCAN variant trained on pedestrian and cyclist classes.
Depending on the experiment between 40 and 60 percent of FPs are due to ghost objects.
Leading to the conclusion that in scenarios where ghost objects occur, they do in fact cause FPs detections.

\begin{table}
	\caption{Percentage of FPs due to different objects. Evaluation done for IoU threshold of 0.3. \emph{BG} stands for background and \emph{intra cls} is confusion of pedestrians with cyclists and vise versa. Second half are the evaluation for pedestrain and object combined.}
	\label{tab:confusion}
	\centering
	\resizebox{\columnwidth}{!}{
		\begin{tabular}{@{}rcccccc@{}}
			\toprule
			\textbf{Method} & \textbf{BG} & \textbf{Intra Cls} & \textbf{MP-12} & \textbf{MP-22} & \textbf{MP-23} & \textbf{OMP} \\ \midrule
			           SGPN &  $39.43\%$  &     $25.13\%$      &   $18.65\%$    &   $12.80\%$    &    $2.89\%$    &   $1.79\%$   \\
			         DBSCAN &  $29.81\%$  &     $11.53\%$      &   $36.57\%$    &   $11.10\%$    &    $6.76\%$    &   $4.28\%$   \\ \midrule
			           SGPN &  $60.42\%$  &                    &   $20.18\%$    &   $13.79\%$    &    $4.77\%$    &   $0.82\%$   \\
			         DBSCAN &  $57.57\%$  &                    &   $25.71\%$    &   $12.14\%$    &    $4.14\%$    &   $0.41\%$   \\ \bottomrule
		\end{tabular}
	}
	\vspace{-5mm}
\end{table} % \label{tab:confusion}

\section{Conclusion}
This article extends an existing dataset with novel and detailed annotations, allowing for a first in-depth analysis of different multi-path occurrences.
Both, the formation and the annotation, of more intricate reflection types are described in detail.
Two baseline methods are provided as an example for ghost detection on an instance level.
In the process, we show how to build a new end to end instance segmentation network for automotive radar data based on the idea behind SGPN.
The SGPN variant performs very well for multi-path detection, while the extended PointNet++ approach better suppresses false positives on these reflection types.
With this dataset, we hope to encourage other researchers to join us in advancing the field of data-driven radar multi-path detection.

\addtolength{\textheight}{-12cm}  % This command serves to balance the column lengths

%%%%%%%%%%%%%%%%%%%%%%%%%%%%%%%%%%%%%%%%%%%%%%%%%%%%%%%%%%%%%%%%%%%%%%%%%%%%%%%%

\bibliographystyle{IEEEtran}
\bibliography{IEEEabrv,bib}
\end{document}